\documentclass{article}

\usepackage[T1]{fontenc}
\usepackage[utf8]{inputenc}
\usepackage[a4paper,margin=1in]{geometry}
\usepackage{lmodern}
\usepackage{microtype}
\usepackage{amsmath,amssymb,amsfonts}
\usepackage{booktabs}
\usepackage{graphicx}
\usepackage{xcolor}
\usepackage{hyperref}
\usepackage{enumitem}
\usepackage{natbib}
\usepackage{array}
\usepackage{siunitx}
\usepackage{placeins}

\hypersetup{
    colorlinks=true,
    linkcolor=blue,
    citecolor=blue,
    urlcolor=blue
}

\title{Agentic Control in Variational Language Models}

\author{%
Yves Ruffenach\\
Conservatoire National des Arts et M\'etiers\\
Strasbourg, France\\
\texttt{yves@ruffenach.net}
}

\date{}

\begin{document}

\maketitle

\begin{abstract}
We study whether a variational language model can support a minimal and measurable form of agentic control grounded in its own internal evidence. Our model combines local variational hidden computation (\textsc{EVE}), a homeostatic latent regulator, structurally aware checkpoint retention, and a calibrated uncertainty-aware controller operating on top of the retained model. Rather than treating uncertainty as a passive diagnostic measured after prediction, we treat it as an operational signal that can regulate training, support checkpoint retention, and guide inference-time intervention. The resulting framework is deliberately focused. It studies a closed-loop form of internal control in which structural and predictive signals become actionable. Empirically, the variational backbone improves over a matched deterministic reference on the language-modeling task while also exhibiting a richer and more usable uncertainty profile. On top of this backbone, the calibrated controller remains active, uses multiple actions under a full agentic evaluation, and yields a positive quality--cost trade-off. These results support a precise claim: internal uncertainty can serve not only as a descriptive property of a variational language model, but also as a practical control interface for regulation, checkpoint retention, and minimal agentic routing.
\end{abstract}

\section{Introduction}

Large language models are now judged not only by predictive quality, but also by how they behave under uncertainty.
This shift has made \emph{agentic behavior} a central topic.
Most recent work studies agency primarily through external mechanisms such as tool use, retrieval, planning, or multi-step action policies layered on top of an already trained model.
That line of work is important.
It also raises a complementary question:
can a language model support a minimal and measurable form of agentic control \emph{from within}, grounded in its own internal evidence?

This paper studies that question in a focused setting.
We ask whether a variational language model can sustain a compact control loop in which internal uncertainty becomes actionable.
Our approach combines four elements:
a variational language-model backbone with local stochastic hidden computation,
a homeostatic regulator that keeps the latent regime in a healthy operating range during training,
a structurally aware checkpoint-retention rule,
and a calibrated uncertainty-aware controller that selects an inference-time action on top of the retained model.
The resulting notion of agency is intentionally focused.
It is also concrete, testable, and measurable.

The central idea of the paper is simple.
Uncertainty becomes more useful when it is treated not only as a descriptive quantity, but also as an operational signal inside a control pipeline.
In our setting, uncertainty is used at three points.
It helps regulate the latent regime during training.
It helps retain a model state that remains structurally meaningful.
It then helps guide inference-time intervention once a final checkpoint has been retained.
In this formulation, uncertainty is not only reported.
It is used to regulate, retain, and route.

This perspective is especially natural for variational hidden computation.
When hidden computation is purely deterministic, post hoc confidence estimates can still be informative, but they expose only part of the model state.
By contrast, local variational hidden units expose richer internal evidence, including KL activity, latent-energy statistics, predictive entropy, mutual information, and disagreement across stochastic passes.
This gives the model a dual role.
It remains a predictor, and it also becomes a source of internal evidence on which a minimal controller can operate.

A central design choice of the paper is to separate three roles that are often blended together.
The \emph{homeostatic regulator} acts during optimization and keeps the latent regime within a viable operating range.
The \emph{checkpoint-retention rule} acts after candidate states have been observed and keeps a model that remains acceptable both on the task and in its internal structural profile.
The \emph{uncertainty-aware controller} acts afterward, at inference time, on top of the retained model.
This separation gives the pipeline a clear order:
regulation first, retention second, calibrated routing third.

We evaluate this idea in controlled next-token language modeling with a frozen GPT-2 embedding front end and a matched deterministic reference.
At a high level, the empirical picture is clear.
The variational backbone improves over the deterministic reference on the main language-modeling metrics while also exposing a richer uncertainty profile.
On top of the retained \textsc{EVE} checkpoint, the calibrated controller remains active under a full multi-action protocol with positive utility and broad coverage.
Together, these results support the main claim of the paper:
a variational language model can sustain a minimal closed-loop decision process in which calibrated internal uncertainty affects what the system does next.

The contribution of the paper is a controlled demonstration that agentic behavior can be studied at the level of \emph{internal model regulation and calibrated intervention}.
More specifically, the paper makes four contributions.

\begin{itemize}[leftmargin=1.5em]
    \item It introduces a variational language-modeling framework in which local stochastic hidden computation produces structurally observable internal signals rather than only final predictions.
    \item It proposes a homeostatic latent regulator that preserves a viable stochastic regime during training and maintains useful latent activity.
    \item It defines a structurally aware checkpoint-retention rule that keeps model states on the basis of both task quality and internal latent meaningfulness.
    \item It shows that a calibrated uncertainty-aware controller built on the retained variational backbone can implement a minimal and measurable form of agentic control at inference time.
\end{itemize}

The rest of the paper follows this logic directly.
Section~\ref{sec:method} presents the model, the latent regulator, the internal signals, the checkpoint-retention rule, the calibrated controller, and the evaluation protocol.
Section~\ref{sec:experiments} reports the empirical results.
The final sections discuss the scope of the findings and summarize the main conclusion.

\paragraph{Minimal agentic loop.}
In this paper, agentic behavior is defined in a deliberately focused and measurable way.
The system reads internal evidence, maps that evidence to a decision, applies a stronger intervention when needed, and operates within a broader pipeline that also includes structurally aware checkpoint retention.
This gives the control logic a clear and operational form.

\begin{figure}[htbp]
\centering
\includegraphics[width=0.9\linewidth]{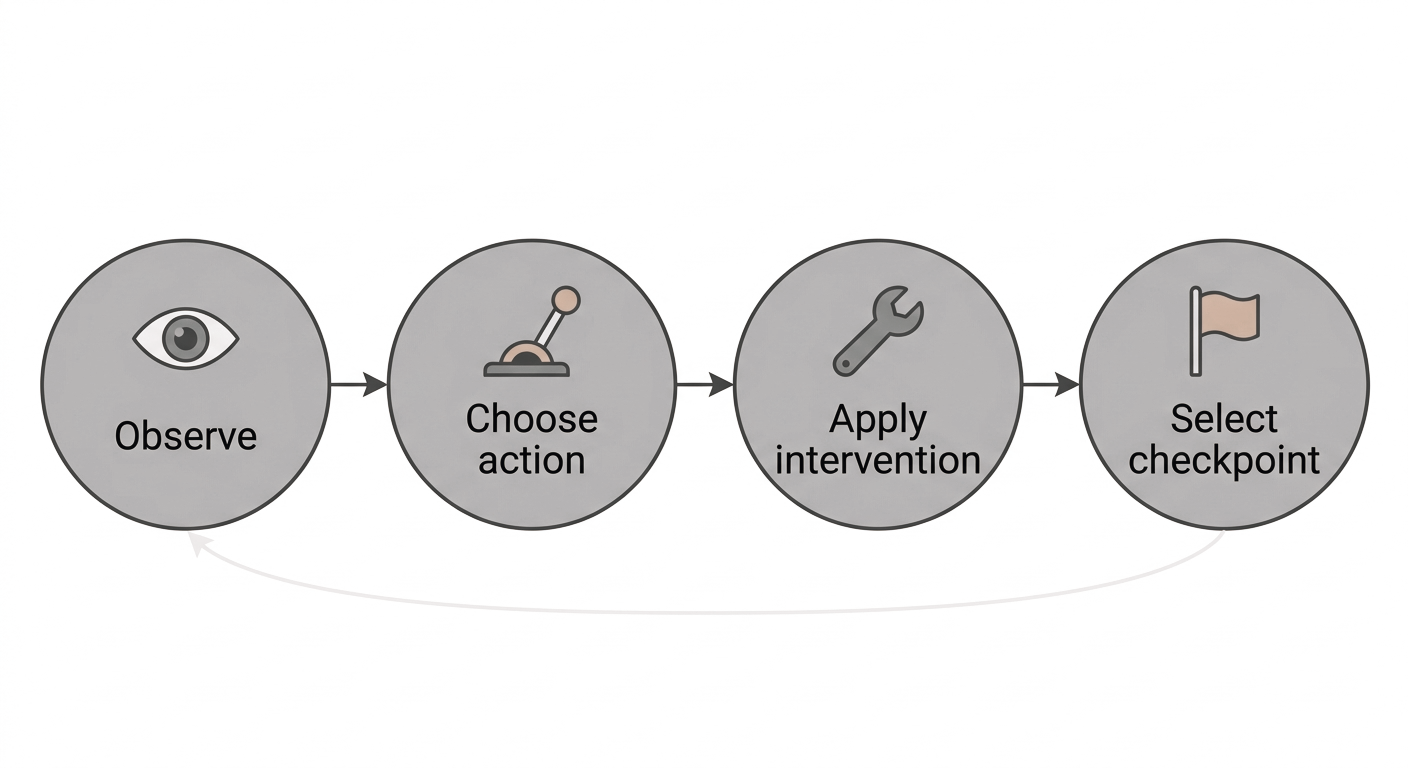}
\caption{
A minimal agentic loop.
Internal evidence is observed, mapped to an action, and followed by stronger intervention when required.
In the full method, this loop is embedded in a broader pipeline that also includes structurally aware checkpoint retention.
}
\label{fig:minimal_agentic_loop}
\end{figure}

\begin{figure}[htbp]
\centering
\includegraphics[width=0.9\linewidth]{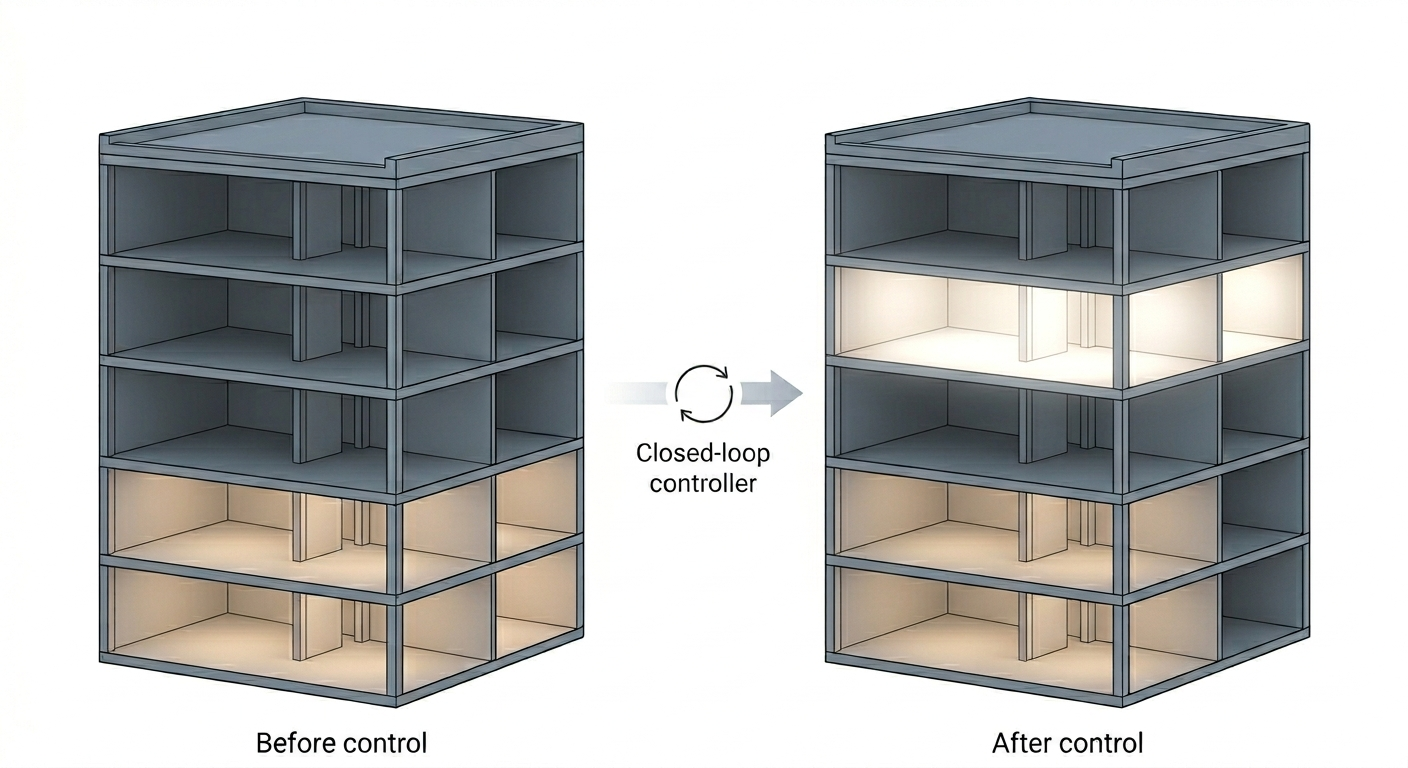}
\caption{
A building metaphor for underused and recruited depth.
Before control, the model learns but deeper floors remain dark.
After closed-loop intervention, a deeper floor becomes active.
}
\label{fig:building_deep_recruitment}
\end{figure}

\paragraph{Visual intuition.}
A simple way to understand our setting is to view the model as a multi-story building.
Task learning may already be present in the lower floors, while deeper floors remain structurally inactive.
The role of the controller is to recruit part of this unused depth without changing the overall task.
Figure~\ref{fig:building_deep_recruitment} illustrates this idea.

\FloatBarrier

\section{Related Work}

\subsection{Variational and Bayesian modeling for language-model uncertainty}

Variational latent-variable models and Bayesian neural networks provide the main probabilistic background for our setting.
The variational autoencoder literature established practical amortized inference for continuous latent variables, while Bayesian neural-network methods showed how uncertainty can be embedded directly into model computation rather than added only at decision time \citep{kingma2014autoencoding, blundell2015weight}.

In modern deep learning, uncertainty is often introduced through weight-level or output-level approximations such as Monte Carlo dropout and deep ensembles \citep{gal2016dropout, lakshminarayanan2017deepensembles}.
Transformer-specific variants follow the same general direction.
BayesFormer extends Bayesian dropout ideas to Transformer architectures, and Bayesian Transformer language models place posterior distributions over broader model components such as attention, feed-forward, and embedding layers \citep{sankararaman2022bayesformer, xue2021bayesiantransformerlm}.

These works provide an important foundation.
They show that uncertainty can be integrated into deep language modeling in a principled way.
Our starting point is close to this literature in spirit, but different in placement.
Rather than introducing uncertainty mainly at the level of weights, dropout masks, or output aggregation, we place stochastic latent computation directly inside local hidden units.
This makes internal latent activity itself a first-class object of analysis.

Our work also builds on prior results showing that local variational neurons can be integrated into language-model hidden computation while preserving strong predictive behavior and exposing informative uncertainty signals \citep{ruffenach2026variationalneurons}.
The present paper extends that line in a more specific direction.
The question is no longer only whether local variational hidden computation yields useful uncertainty signals, but whether those signals can support regulation, checkpoint retention, and calibrated inference-time intervention inside one coherent control pipeline.

\subsection{Uncertainty, calibration, and selective prediction}

A second line of work studies how predictive uncertainty should be measured and used.
Predictive entropy, conditional entropy, mutual information, disagreement across stochastic forward passes, and calibration metrics such as expected calibration error are standard tools in Bayesian deep learning and uncertainty evaluation \citep{gal2016dropout, lakshminarayanan2017deepensembles, guo2017calibration}.
These quantities are widely used to assess reliability, detect overconfidence, and compare uncertainty estimators.

A closely related literature turns uncertainty into selective action.
Selective classification and reject-option methods formalized the idea that a predictor can improve reliability by abstaining on uncertain cases, and integrated approaches such as SelectiveNet made the selection rule part of the model itself \citep{geifman2017selectiveclassification, geifman2019selectivenet}.
More recent work has brought the same concern to language models, arguing that large language models should be trained to know when they do not know rather than relying on confidence heuristics alone \citep{kapoor2024llmsmustbetaught}.

Our work is close to this literature in its use of uncertainty as a decision variable.
The difference lies in the scope of the decision.
We do not study uncertainty only as a confidence report, an abstention score, or a post hoc reliability diagnostic.
Instead, we treat calibrated uncertainty as part of a minimal control interface.
The question is not only whether a prediction should be trusted, but whether the current predictive state calls for direct prediction, stronger stochastic deliberation, retrieval-style support, or abstention.
In this sense, uncertainty is used not only to describe reliability, but also to organize action.

\subsection{Internal regulation and structurally aware model retention}

A third relevant line of work concerns internal regulation and model-state retention.
Many training pipelines remain effectively task-only at the checkpoint level: the retained state is the one that optimizes a validation metric, regardless of whether the internal regime that produced that score remains stable, meaningful, or useful for downstream control.
That choice is often sufficient when a model is used only as a predictor.
It is less sufficient when the same model is also expected to support uncertainty-based intervention.

The present paper follows a different intuition.
Checkpoint retention can itself be part of the control story.
A retained state should remain acceptable on the task, but it should also preserve an internal regime that stays active, regulated, and interpretable under a fixed evaluation view.
This makes retention a bridge between variational training and later uncertainty-aware decision making.

More broadly, this position is consistent with a familiar idea in adaptive systems:
internal regulation and external decision should be distinguished rather than merged.
Our paper follows that logic explicitly.
Latent regulation during training, checkpoint retention after training, and calibrated routing at inference time are treated as separate roles inside one ordered pipeline.

\subsection{Agentic AI, retrieval, and decision control}

A fourth line of work frames agentic language models around external action.
Retrieval-augmented generation augments a language model with non-parametric memory, MRKL systems emphasize modular external reasoning and knowledge components, ReAct interleaves reasoning traces with environment-facing actions, and Toolformer trains language models to decide when and how to call external APIs \citep{lewis2020rag, karpas2022mrkl, yao2023react, schick2023toolformer}.

This literature is highly relevant as a point of comparison.
In these systems, the central decision is typically whether to retrieve, call a tool, query an environment, or delegate part of the task to another module.
Our paper studies a different layer of the problem.
The controller does not aim to establish a full external-action agent architecture.
Instead, it operates on top of a variational language model whose internal state is already observable and calibrated.
The relevant decision is whether the current predictive situation should be handled directly or whether a stronger intervention is warranted.

This makes the notion of agency studied here intentionally focused.
The goal is not a general-purpose tool-using agent.
The goal is to show that a language model can read internal evidence, preserve a meaningful latent regime through training-time regulation and checkpoint retention, and then use calibrated uncertainty to decide what should happen next at inference time.

\paragraph{Position of this work.}
The closest conceptual neighbors of the present paper lie at the intersection of three traditions:
variational and Bayesian models with measurable internal uncertainty,
calibration and selective-prediction methods that turn uncertainty into action,
and agentic systems that make state-dependent decisions rather than issuing unconditional predictions.
Our contribution is to connect these traditions inside one variational language-model setting.
Internal uncertainty is not only estimated after the fact.
It is regulated during training, preserved through structurally aware checkpoint retention, and then used as a calibrated control signal for minimal agentic routing.

\section{Method}
\label{sec:method}

\subsection{Overview}

We study a variational next-token language model with a minimal internal control loop.
The method has four parts:
\begin{enumerate}[leftmargin=1.5em]
    \item a frozen GPT-2 embedding front end with trainable hidden computation after the embedding layer,
    \item local variational hidden computation with homeostatic latent regulation,
    \item a structurally aware checkpoint-retention rule,
    \item and a calibrated uncertainty-aware controller applied to the retained model.
\end{enumerate}

The design is intentionally compact.
The system first learns under latent-state regulation.
It then retains a checkpoint that is both task-safe and structurally usable.
Finally, it maps calibrated uncertainty to an inference-time action.
This defines a focused and measurable form of agentic control grounded in internal evidence.

Empirically, we evaluate this control logic in a full multi-action setting designed to test active uncertainty-aware routing under an explicit quality--cost trade-off.

\subsection{Task setting and controlled comparison}

The task is next-token language modeling on prompt--story examples.
Each example is tokenized with GPT-2.
We use the GPT-2 input embedding matrix as a fixed front end.
More precisely, we load the GPT-2 tokenizer and backbone, extract the input embedding matrix, and freeze it through
\texttt{nn.Embedding.from\_pretrained(..., freeze=True)}.
The learned computation therefore begins after the token embedding layer.

This point is central to the comparison.
The study isolates the effect of the hidden computation by keeping the GPT-2 front end fixed.
It then evaluates whether replacing deterministic hidden computation with variational hidden computation changes both predictive quality and the usefulness of internal uncertainty.

We compare two matched model families:
\begin{itemize}[leftmargin=1.5em]
    \item \textsc{EVE}, which uses local variational hidden computation,
    \item \textsc{DET}, which keeps the same frozen GPT-2 embedding front end, the same task format, and the same prediction interface, but removes the posterior, prior, and latent sampling path.
\end{itemize}

The comparison is therefore controlled at the level of tokenizer, input embeddings, task, and prediction interface.
The main difference lies in the hidden computation and in the availability of internal stochastic evidence.

\subsection{Data and preprocessing}

The notebook uses \texttt{RLAIF/WritingPrompts-Filtered} with the training split only.
A fixed fraction of the split is retained, with \texttt{dataset\_fraction = 0.01}.
The selected subset is then partitioned with a fixed validation fraction \texttt{val\_frac = 0.20}.

Each raw example is converted into a prompt--story pair.
The tokenizer is GPT-2.
The prompt is truncated to at most \texttt{max\_prompt\_tokens = 96} tokens.
The story is truncated to at most \texttt{max\_story\_tokens = 192} tokens.
The hidden module reads a fixed context window of length \texttt{context\_len = 24}.
The target stride is \texttt{target\_stride = 2}.
Examples with stories shorter than \texttt{min\_story\_tokens = 32} are filtered out.

This yields a fixed and reproducible next-token evaluation setup.

\subsection{Frozen front end and prediction head}

Let $x_{1:T}$ denote the tokenized prompt context.
The frozen front end maps each token to its GPT-2 embedding.
These embeddings are concatenated over the context window and passed to the trainable hidden module.
The prediction head then maps the hidden representation back to the vocabulary space through a token-prediction head tied to the frozen embedding table.

In the reported setup, the GPT-2 input embeddings remain fixed throughout training and evaluation.
The trainable computation begins after the embedding layer and includes the hidden module together with the output-side trainable parameters allowed by the tied prediction setup.
The experiment therefore keeps the GPT-2 front end fixed while learning the downstream computation needed for the controlled comparison.

\subsection{Variational hidden computation}

The variational model, denoted \textsc{EVE}, uses local latent variables inside hidden computation.
For a hidden unit or hidden block, the model defines
\begin{align}
q_{\phi}(z \mid x, h),
\qquad
p_{\psi}(z \mid h),
\qquad
z = \mu_q + \sigma_q \odot \varepsilon,
\qquad
\varepsilon \sim \mathcal{N}(0, I).
\end{align}

In the reported experiments, local variational hidden computation is instantiated inside a transformer-style hidden stack.
The default configuration uses:
\begin{itemize}[leftmargin=1.5em]
    \item \texttt{use\_variational\_transformer = True},
    \item \texttt{transformer\_num\_heads = 12},
    \item \texttt{variational\_mlp\_layers = 3},
    \item \texttt{variational\_mlp\_hidden = 1024},
    \item \texttt{latent\_mc\_samples\_train = 4},
    \item \texttt{latent\_mc\_samples\_eval = 12}.
\end{itemize}

The deterministic baseline \textsc{DET} preserves the same task frame and prediction interface but removes the latent posterior, latent prior, and stochastic sampling path.

\subsection{Training objective}

Let $\mathcal{L}_{\mathrm{CE}}$ denote the next-token cross-entropy.
For \textsc{DET}, the optimized loss is simply
\begin{align}
\mathcal{L}_{\mathrm{DET}} = \mathcal{L}_{\mathrm{CE}}.
\end{align}

For \textsc{EVE}, the notebook optimizes
\begin{align}
\mathcal{L}_{\mathrm{EVE}}
=
\mathcal{L}_{\mathrm{CE}}
+
\mathcal{L}_{\mathrm{regKL}}
+
\lambda_{\mathrm{local}} \mathcal{L}_{\mathrm{local}}
+
\lambda_{\mathrm{band}} \mathcal{L}_{\mathrm{band}},
\end{align}
where:
\begin{itemize}[leftmargin=1.5em]
    \item $\mathcal{L}_{\mathrm{CE}}$ is the task loss,
    \item $\mathcal{L}_{\mathrm{regKL}}$ is the regularized latent KL contribution,
    \item $\mathcal{L}_{\mathrm{local}}$ is the local reconstruction term,
    \item $\mathcal{L}_{\mathrm{band}}$ is the homeostatic band penalty.
\end{itemize}

In code form, this is:
\begin{align}
\texttt{loss}
=
\texttt{ce}
+
\texttt{reg\_kl}
+
\texttt{lam\_local\_recon} \times \texttt{local\_recon}
+
\texttt{lam\_band} \times \texttt{band}.
\end{align}

The default regularization scales in the notebook are
\texttt{lam\_local\_recon = 3e-2},
\texttt{lam\_band = 4e-3},
and \texttt{mu2\_target = 0.10}.

Throughout the paper, \emph{Loss} denotes the full optimized objective for \textsc{EVE} and coincides with \emph{CE} for \textsc{DET}.

\subsection{Homeostatic latent regulation}

The latent regime is regulated during training by a homeostatic autopilot.
Its role is to keep the latent state in a viable operating zone.
The notebook uses \texttt{control\_mode = "homeo"} together with
\texttt{use\_beta\_thermostat = True}
and
\texttt{use\_neuron\_regulator = True}.

The regulator tracks internal quantities tied to latent activation and stability, including:
\begin{itemize}[leftmargin=1.5em]
    \item latent energy,
    \item band occupancy,
    \item the fraction of units that are too low,
    \item the fraction of units that are too high,
    \item local reconstruction activity,
    \item and KL-related quantities.
\end{itemize}

Its goal is simple.
The latent regime should stay active enough to remain informative and controlled enough to remain usable.
In addition to the base band penalty, the search procedure also explores the strength of the high-activity penalty component, denoted here as $\lambda_{\text{band,high}}$.

\paragraph{Visual intuition.}
The homeostatic controller acts like a thermostat.
If the regime becomes too cold, stochastic structure fades.
If it becomes too hot, latent activity becomes less reliable.
The regulator keeps the model in a bounded middle zone.

\begin{figure}[htbp]
\centering
\includegraphics[width=0.9 \linewidth]{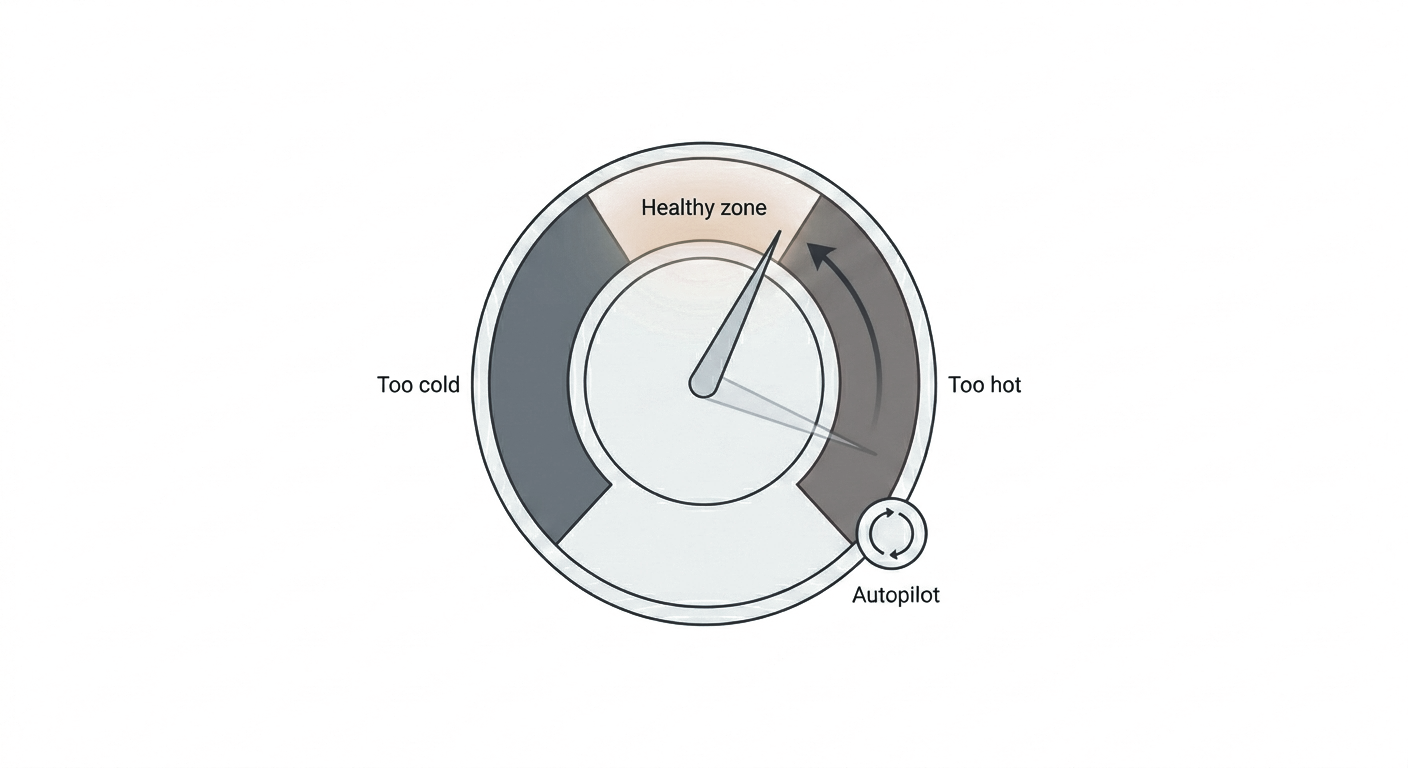}
\caption{
A thermostat metaphor for homeostatic latent regulation.
The autopilot keeps the latent regime in a healthy zone: active, stable, and controlled.
}
\label{fig:thermostat_autopilot}
\end{figure}

\FloatBarrier

\subsection{Internal signals}

The method reads both structural and predictive signals from the model.

\paragraph{Structural signals.}
The main structural quantities used in the notebook are:
\begin{itemize}[leftmargin=1.5em]
    \item \textbf{KL activity}: the latent KL contribution under the evaluation view,
    \item \textbf{Local reconstruction}: the local latent reconstruction proxy used in the loss and in selection,
    \item \textbf{Latent energy} $\mu^2$: the mean squared latent mean,
    \begin{align}
    \mu^2 = \frac{1}{U}\sum_{u=1}^{U}\mu_u^2,
    \end{align}
    \item \textbf{Band occupancy}: the fraction of units inside the target range,
    \item \textbf{Frac.\ Low}: the fraction of units below the lower target bound,
    \item \textbf{Frac.\ High}: the fraction of units above the upper target bound.
\end{itemize}

In particular, the notebook uses \texttt{frac\_too\_high} and \texttt{mu2\_mean\_eval} inside checkpoint retention.

\paragraph{Predictive signals.}
For a given prompt $x$, the notebook draws repeated stochastic forward passes and computes:
\begin{itemize}[leftmargin=1.5em]
    \item \textbf{Predictive entropy},
    \begin{align}
    H(x) = - \sum_y \bar{p}(y \mid x)\log \bar{p}(y \mid x),
    \end{align}
    where $\bar{p}$ is the Monte Carlo predictive mean,
    \item \textbf{Conditional entropy},
    \item \textbf{Mutual information},
    \begin{align}
    \mathrm{MI}(x)
    =
    H(x)
    -
    \frac{1}{M}\sum_{m=1}^{M}
    \left(
    -\sum_y p^{(m)}(y \mid x)\log p^{(m)}(y \mid x)
    \right),
    \end{align}
    \item \textbf{Epistemic uncertainty summary}, abbreviated \textbf{Epi.} in the results tables, reported by the notebook as an auxiliary stochastic-disagreement summary,
    \item \textbf{Top-1 flip rate under Monte Carlo sampling}, denoted \texttt{top1\_flip\_rate\_mc},
    \item \textbf{Confidence}, defined from the predictive distribution.
\end{itemize}

\paragraph{Applicability to \textsc{DET}.}
Task metrics such as CE, NLL, perplexity, accuracy, and calibration metrics apply to both model families.
Purely variational quantities such as latent KL are reported for \textsc{EVE}, while structural activation summaries for \textsc{DET} are reported through an explicit deterministic analogue when such a counterpart is defined.

\subsection{Canonical evaluation view}

All checkpoint comparisons are made under a fixed canonical evaluation view.
This view keeps the same tokenizer, frozen embedding front end, preprocessing, context length, and Monte Carlo evaluation protocol across all candidate checkpoints.
No training-time adaptation is allowed inside this view.

Checkpoint retention and uncertainty-based routing therefore rely on the same observation protocol.

\subsection{Separation of regulation, retention, and control}

The method separates three roles.

First, the \textbf{homeostatic regulator} acts during training.
It keeps the latent regime inside a usable operating zone.

Second, the \textbf{checkpoint-retention rule} acts after candidate states have been observed.
It keeps a checkpoint that is both task-safe and structurally meaningful.

Third, the \textbf{uncertainty-aware controller} acts at inference time on the retained model.
It maps the current uncertainty state to an action.

This separation gives the method a clear order:
regulation first, retention second, calibrated control third.

\subsection{Structurally aware checkpoint retention}

Checkpoint retention is procedural in the notebook.
It is not based on task loss alone.

\paragraph{Task-safe filter.}
For a set of candidate epoch records, let \texttt{best\_val\_ce} be the best observed validation CE and let \texttt{best\_val\_local} be the best observed validation local reconstruction.
A candidate is marked \texttt{task\_safe} if:
\begin{align}
\texttt{ce} &\le \texttt{best\_val\_ce} + \texttt{selection\_ce\_tolerance}, \\
\texttt{local\_recon} &\ge \texttt{selection\_local\_ratio} \times \texttt{best\_val\_local},
\end{align}
with defaults
\texttt{selection\_ce\_tolerance = 0.03}
and
\texttt{selection\_local\_ratio = 0.90},
and with the additional requirement that the record is finite.

\paragraph{Structural sort key.}
Candidates are then ranked with the sort key
\begin{align}
\Big(
&\mathbb{1}[\neg \texttt{task\_safe}],
\;
\texttt{frac\_too\_high},
\;
|\texttt{mu2\_mean\_eval} - \texttt{mu2\_target}|,
\;
-\texttt{acc},
\;
\texttt{ce},
\;
-\texttt{local\_recon}
\Big).
\end{align}
This means that the method prefers, in order:
\begin{enumerate}[leftmargin=1.5em]
    \item task-safe candidates,
    \item lower \texttt{frac\_too\_high},
    \item smaller distance to the target latent energy \texttt{mu2\_target},
    \item higher accuracy,
    \item lower CE,
    \item higher local reconstruction.
\end{enumerate}

\paragraph{Final EVE candidate across seeds.}
After the per-run retention step, the notebook selects the final \textsc{EVE} candidate across final seed-locked candidates with
\begin{align}
(\texttt{val\_ce}, -\texttt{val\_acc}, \texttt{epoch}),
\end{align}
that is, lower validation CE first, then higher validation accuracy, then earlier epoch.
The stored selection reason is \texttt{best\_observed\_final\_validation}.

This final cross-seed choice is made only among checkpoints that have already passed the within-run structurally aware retention step.
Structural constraints therefore act first, while the final seed choice is validation-driven within that admissible set.
This makes checkpoint retention explicit and reproducible.

\subsection{Unified uncertainty score}

The controller operates on a single uncertainty score computed from three primary metrics:
predictive entropy,
mutual information,
and top-1 flip rate under Monte Carlo sampling.

For an input $x$, the notebook first normalizes each component with fixed caps:
\begin{align}
e(x) &= \mathrm{clip}\!\left(\frac{\texttt{predictive\_entropy}(x)}{\texttt{uq\_norm\_entropy}}, 0, 1\right), \\
m(x) &= \mathrm{clip}\!\left(\frac{\texttt{mutual\_information}(x)}{\texttt{uq\_norm\_mi}}, 0, 1\right), \\
f(x) &= \mathrm{clip}\!\left(\frac{\texttt{top1\_flip\_rate\_mc}(x)}{\texttt{uq\_norm\_flip}}, 0, 1\right).
\end{align}

The default notebook values are:
\begin{align}
\texttt{uq\_norm\_entropy} &= 3.5, \\
\texttt{uq\_norm\_mi} &= 0.50, \\
\texttt{uq\_norm\_flip} &= 0.35.
\end{align}

The uncertainty score is then computed as a weighted mean:
\begin{align}
U(x)
=
\frac{
w_e e(x) + w_m m(x) + w_f f(x)
}{
w_e + w_m + w_f
},
\end{align}
with default notebook weights
\begin{align}
w_e = 0.40,
\qquad
w_m = 0.40,
\qquad
w_f = 0.20.
\end{align}

The notebook then adds a small confidence penalty:
if \texttt{confidence} $< \texttt{agent\_answer\_min\_conf}$ with default
\texttt{agent\_answer\_min\_conf = 0.20},
the score is increased by $0.10$ and clipped to $1.0$.
This keeps the policy conservative when predictive confidence is too low.

\subsection{Calibration of routing thresholds}

Raw uncertainty is not yet a policy.
The notebook therefore calibrates routing thresholds on held-out validation rows.

\paragraph{Basic threshold calibration.}
For a list of validation prompts, the model computes one uncertainty score per prompt.
Given quantiles
\texttt{agent\_threshold\_quantiles = (0.50, 0.75, 0.90)},
the notebook sets:
\begin{align}
\texttt{uq\_green} &= Q_{0.50}(U), \\
\texttt{uq\_orange} &= Q_{0.75}(U), \\
\texttt{uq\_red} &= Q_{0.90}(U),
\end{align}
where $Q_q(U)$ denotes the empirical quantile of the score distribution on calibration rows.

\paragraph{Full multi-action calibration.}
For the full controller, the notebook searches threshold triplets over
\texttt{agentic\_full\_threshold\_quantile\_grid}.
Each triplet defines
\texttt{uq\_green},
\texttt{uq\_orange},
and
\texttt{uq\_red}
as score quantiles.
The selected triplet maximizes a penalized proxy objective that rewards utility and coverage and controls abstain and retrieve rates.

\subsection{Action space and routing policy}

The notebook uses the following low-level actions:
\begin{itemize}[leftmargin=1.5em]
    \item \texttt{answer},
    \item \texttt{deliberate\_more},
    \item \texttt{retrieve\_or\_resample},
    \item \texttt{abstain\_or\_escalate}.
\end{itemize}

The basic score-to-action rule is:
\begin{align}
a(x)=
\begin{cases}
\texttt{answer}, & U(x) < \texttt{uq\_green},\\
\texttt{deliberate\_more}, & \texttt{uq\_green} \le U(x) < \texttt{uq\_orange},\\
\texttt{retrieve\_or\_resample}, & \texttt{uq\_orange} \le U(x) < \texttt{uq\_red},\\
\texttt{abstain\_or\_escalate}, & U(x) \ge \texttt{uq\_red}.
\end{cases}
\end{align}

At the evaluation-report level, these actions are grouped into the more readable categories
\texttt{direct},
\texttt{deliberate},
\texttt{retrieve},
and
\texttt{abstain}.

\paragraph{Exclusive and non-exclusive rates.}
Coverage and abstention are terminal outcomes and are therefore exclusive.
Retrieve, deliberate, and resample are support actions and may overlap in traces and aggregate summaries.

\subsection{Final decision after intervention}

Each action produces a post-action predictive state.
For direct answering, this is the base predictive distribution.
For deliberation or resampling, this is the predictive distribution after additional stochastic computation.
For retrieval-style support, this is the predictive distribution obtained after context augmentation and rerunning the model.

The final token decision is the top prediction of the terminal post-action predictive distribution:
\begin{align}
\hat{y}(x) = \arg\max_y p_{\mathrm{final}}(y \mid x),
\end{align}
unless the controller abstains.

\subsection{Cost model and utility}

The notebook evaluates control under an explicit quality--cost trade-off.
The default action costs are:
\begin{align}
\texttt{agent\_cost\_answer} &= 1.0, \\
\texttt{agent\_cost\_resample} &= 2.0, \\
\texttt{agent\_cost\_retrieve\_resample} &= 3.0, \\
\texttt{agent\_cost\_abstain} &= 0.5.
\end{align}

For the publication summary, mean utility is defined as
\begin{align}
\mathrm{Utility}
=
\mathrm{mean}(\mathrm{overall\_acc})
-
\lambda_{\mathrm{cost}} \cdot \mathrm{mean}(\mathrm{cost}),
\end{align}
with
\begin{align}
\lambda_{\mathrm{cost}} = \texttt{publication\_policy\_cost\_weight} = 0.02.
\end{align}

For the agentic evaluation helpers, the notebook uses the same cost-penalized form with
\texttt{agentic\_ cost\_ weight = 0.02}.
The publication summary uses the term \emph{Utility} for this cost-penalized quality criterion.

When reported, \emph{Mean MC cost} refers to the average stochastic-computation burden induced by repeated Monte Carlo evaluation, reported separately from the coarse action-level mean cost.

\subsection{Why this counts as agentic control}

The method satisfies a minimal and measurable notion of agentic behavior:
\begin{itemize}[leftmargin=1.5em]
    \item the system reads its own internal predictive and structural state,
    \item it stabilizes that state during training,
    \item it retains a model state on the basis of both task and structural evidence,
    \item and it uses calibrated uncertainty to choose how to act at inference time.
\end{itemize}

In this framework, uncertainty is not only descriptive.
It is part of the decision process itself.

\subsection{Training protocol}

The notebook uses:
\begin{itemize}[leftmargin=1.5em]
    \item \texttt{batch\_size = 48},
    \item \texttt{epochs = 3} for the main run,
    \item \texttt{lr\_eve = 2e-4},
    \item \texttt{lr\_det = 2e-4},
    \item \texttt{weight\_decay = 1e-4},
    \item gradient clipping with \texttt{clip\_grad = 1.0}.
\end{itemize}

Model development follows a compact search-and-confirm procedure.
A short refinement stage identifies a promising hyperparameter region.
Promising settings are then confirmed across seeds.
The final retained checkpoint is selected only after evaluation under the structurally aware retention rule defined above.

\subsection{Pipeline summary}

The full procedure is:

\begin{enumerate}[leftmargin=1.5em]
    \item build the prompt--story next-token setup with GPT-2 tokenization and frozen GPT-2 input embeddings,
    \item train candidate \textsc{EVE} models with homeostatic regulation,
    \item evaluate candidate checkpoints under the canonical evaluation view,
    \item retain the checkpoint with the structural selection rule,
    \item calibrate uncertainty thresholds on held-out validation rows,
    \item apply the calibrated controller at inference time,
    \item report predictive, structural, calibration, and cost-aware agentic metrics.
\end{enumerate}

\subsection{Evaluation protocol}

We evaluate the method along four complementary axes: predictive quality, uncertainty and calibration, latent-state diagnostics, and full multi-action agentic behavior.

\subsubsection{Predictive quality}

We evaluate predictive quality with cross-entropy, negative log-likelihood, perplexity, and accuracy.
These metrics establish whether the variational backbone remains strong on the base next-token task.

\subsubsection{Uncertainty and calibration}

We evaluate uncertainty with predictive entropy, conditional entropy, mutual information, the auxiliary epistemic-disagreement summary denoted \emph{Epi.}, top-1 Monte Carlo flip rate, and confidence-sensitive threshold calibration.
These metrics determine whether uncertainty is informative enough to support routing.

\subsubsection{Latent-state diagnostics}

We evaluate the health of the latent regime through KL activity, local reconstruction, latent energy, band occupancy, and the fractions of units below or above the target operating range.
These diagnostics show whether the retained checkpoint is both task-valid and structurally usable.

\subsubsection{Agentic evaluation}

The agentic evaluation uses a full multi-action protocol with direct answering, deliberation, retrieval-style support, and abstention.
Its purpose is to test whether calibrated uncertainty can drive a broader action space under an explicit quality--cost trade-off.

\paragraph{Coverage and CE conventions.}
Coverage is the fraction of examples that receive a final answer.
Accepted accuracy is computed on answered examples only.
Accepted CE is computed on answered examples only.
Overall CE is computed over examples with a finite emitted CE.
When abstained examples emit no final predictive distribution, their CE is recorded as non-finite and excluded from overall CE.
In that case, accepted CE and overall CE can coincide by construction, while coverage and utility continue to reflect abstention.

\paragraph{Additional agentic summary.}
We also report \emph{avoided errors versus direct prediction}, defined as the fraction of examples incorrectly handled by the direct policy but correctly handled after routed intervention.

\section{Experiments}
\label{sec:experiments}

The full pipeline produces a clear and coherent result. The final \textsc{EVE} backbone improves over the deterministic reference on the retained validation metrics, exposes usable uncertainty structure, and supports a calibrated multi-action controller with positive downstream utility.

\begin{figure}[htbp]
\centering
\includegraphics[width=0.96\linewidth]{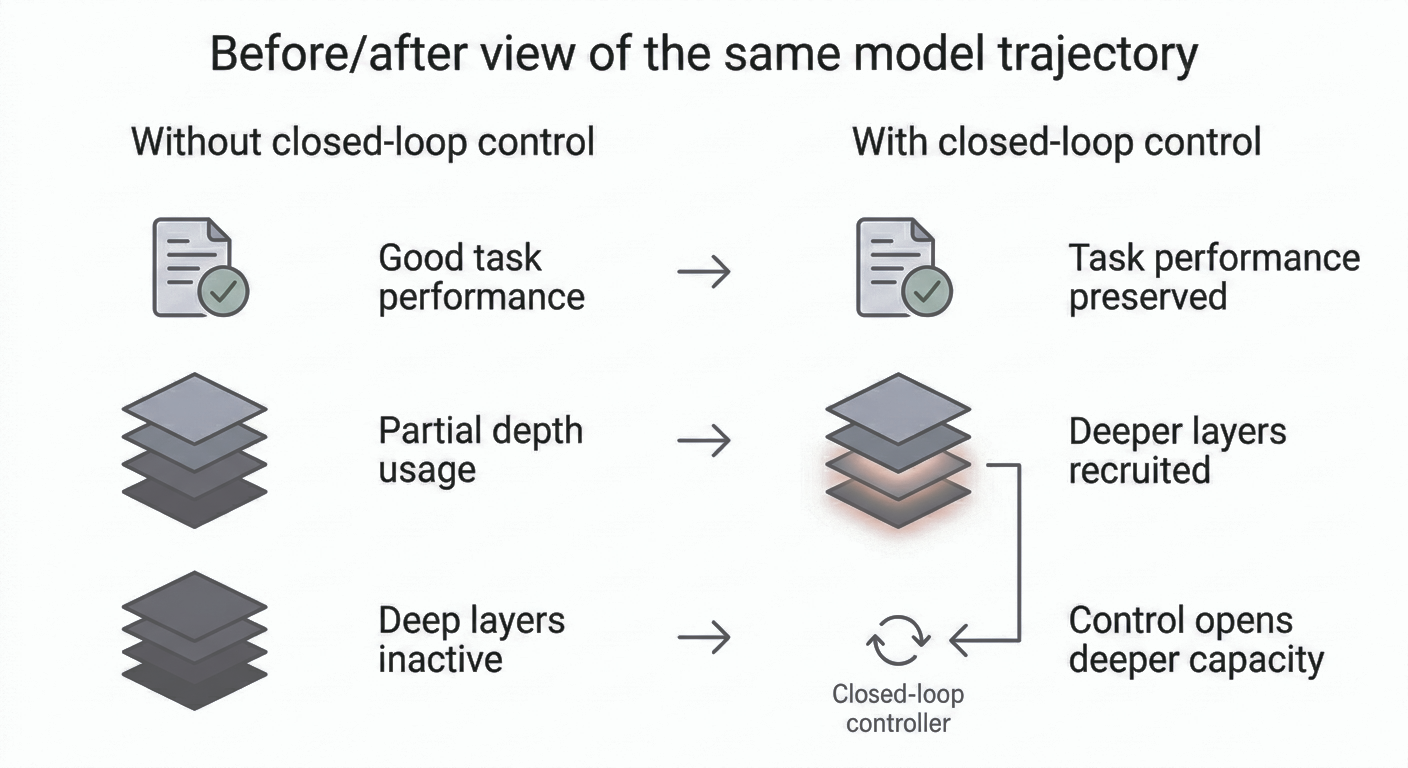}
\caption{
Before/after view of the observed system behavior. The variational backbone supports prediction and exposes internal uncertainty signals that are later used by the calibrated controller.
}
\label{fig:before_after_control}
\end{figure}

\paragraph{Visual summary.}
Figure~\ref{fig:before_after_control} gives a compact view of the observed shift from prediction-only behavior to prediction with calibrated control. It summarizes the empirical behavior reported in this section.

\FloatBarrier

\subsection{Final EVE run}

The retained run combines a strong variational backbone, stable latent activity, and a calibrated controller. The notebook-level validation summary reports \texttt{best\_epoch=2} for both model families. In the later full agentic evaluation, the checkpoint metadata reports \texttt{preferred\_seed=None}, \texttt{selected\_seed=101}, and \texttt{preference\_applied=False}. We therefore use the family-level summary as a coarse validation view and the retained seed-level traces as the authoritative view for the final \textsc{EVE} checkpoint family.

\subsubsection{Final backbone comparison}

Table~\ref{tab:final_backbone_eve_det} reports the final validation comparison between \textsc{EVE} and \textsc{DET}. On this run, \textsc{EVE} achieves lower loss, lower CE, lower perplexity, and higher validation accuracy.

\begin{table}[htbp]
\centering
\caption{Final validation comparison between EVE and DET on the new run.}
\label{tab:final_backbone_eve_det}
\begin{tabular}{lcccccc}
\hline
Model & Loss & CE & PPL & Acc. & KL & Local \\
\hline
EVE & 5.5340 & 5.4341 & 229.09 & 0.1632 & 282.2273 & 0.1309 \\
DET & 5.6035 & 5.6035 & 271.36 & 0.1560 & 0.0000 & 0.0000 \\
\hline
\end{tabular}
\end{table}

Relative to \textsc{DET}, \textsc{EVE} reduces validation CE by $0.1694$ points and perplexity by $42.27$. Validation accuracy increases by $0.0072$ in absolute terms. These results show that the retained variational backbone delivers both stronger predictive quality and richer internal structure.

\begin{figure}[htbp]
\centering
\includegraphics[width=0.9\linewidth]{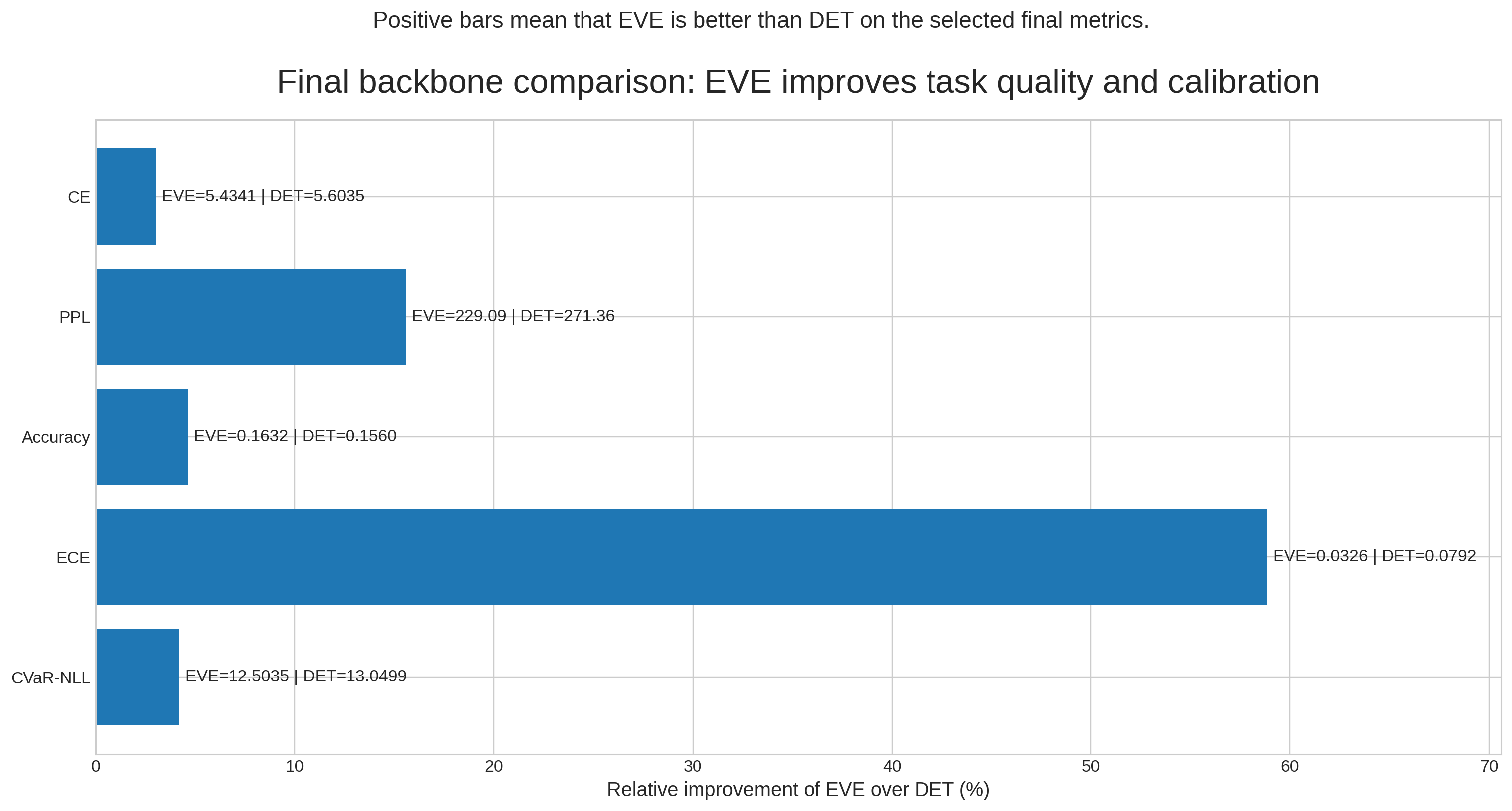}
\caption{
Observed backbone differences between \textsc{EVE} and \textsc{DET} on selected validation metrics. Positive bars indicate an improvement of \textsc{EVE} relative to \textsc{DET}.
}
\label{fig:backbone_relative_gains}
\end{figure}

\FloatBarrier

\subsubsection{Retained seed and final seed-level traces}

Table~\ref{tab:seed_selection_final} reports the final per-seed \textsc{EVE} traces from the retained run. Both seeds continue improving through epoch~3 in validation CE. Seed \texttt{202} reaches the lowest validation CE, and seed \texttt{101} is the one carried into the reported full agentic evaluation.

\begin{table}[htbp]
\centering
\caption{Final EVE seed comparison on the new run.}
\label{tab:seed_selection_final}
\begin{tabular}{lcccccc}
\hline
Seed & Best epoch & Val CE & Val PPL & Val Acc. & Val KL & Val Local \\
\hline
EVE S101 & 3 & 5.4241 & 226.82 & 0.1657 & 255.1250 & 0.1338 \\
EVE S202 & 3 & 5.4161 & 225.00 & 0.1654 & 304.1061 & 0.1404 \\
\hline
\end{tabular}
\end{table}

The CE gap between the two seeds remains small. Seed \texttt{202} provides the strongest validation CE, while seed \texttt{101} stays very close and supports the reported end-to-end agentic evaluation. Together, the two traces confirm a stable and reproducible \textsc{EVE} regime.

\subsubsection{Training dynamics}

\begin{figure}[htbp]
\centering
\includegraphics[width=0.9\linewidth]{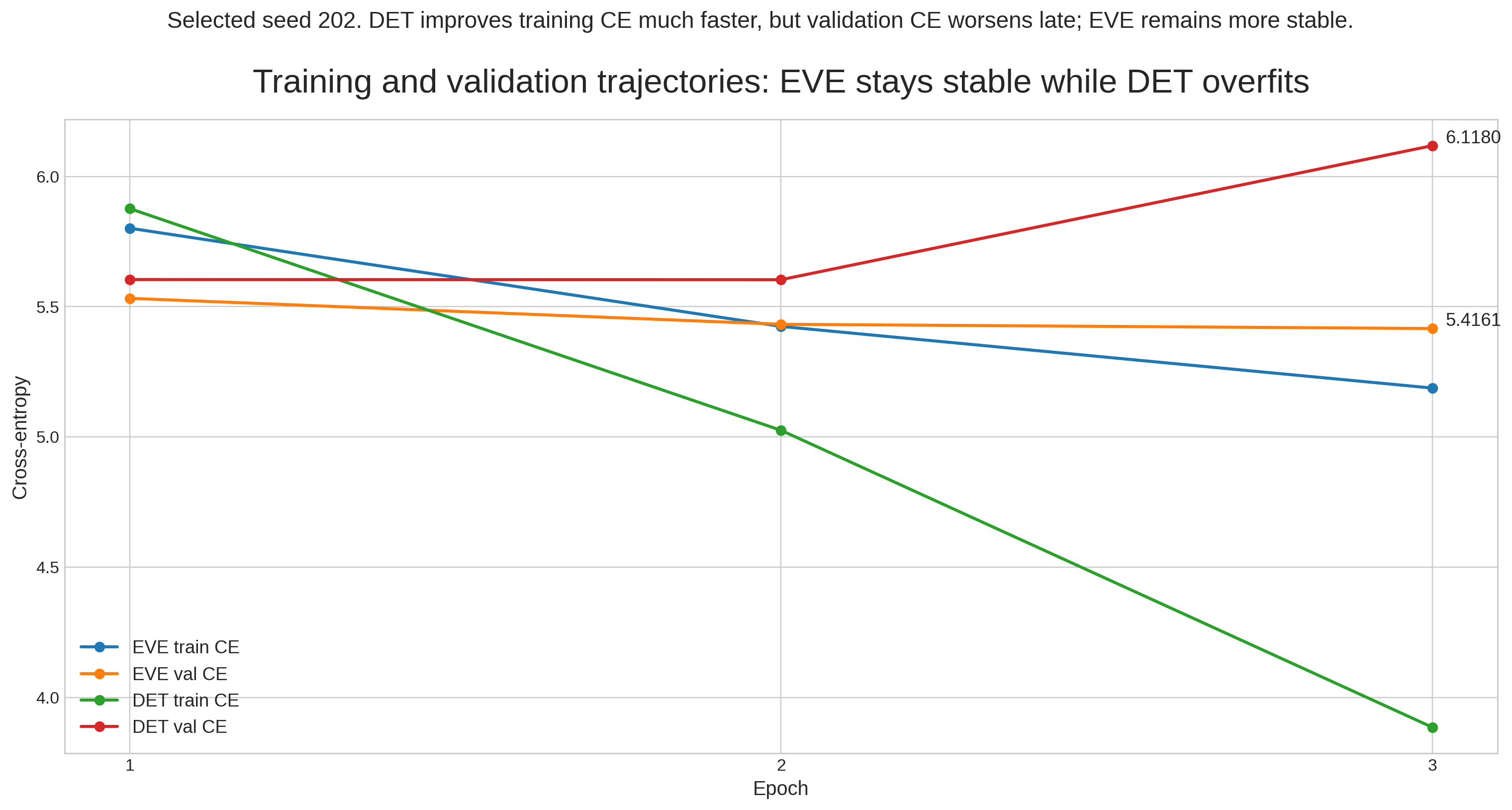}
\caption{
Observed training and validation cross-entropy trajectories for the retained run.
}
\label{fig:train_val_ce_trajectory}
\end{figure}

Figure~\ref{fig:train_val_ce_trajectory} reports the observed training and validation CE trajectories for the retained run and the deterministic reference.

\FloatBarrier

The training dynamics are well structured. \textsc{EVE} improves steadily across epochs and maintains a stable validation profile. The retained \textsc{EVE} traces continue improving through epoch~3, reaching validation CE $5.4241$ for seed \texttt{101} and $5.4161$ for seed \texttt{202}. In the matched reference, the strongest validation regime appears at epoch~2 with validation CE $5.6035$ for seed \texttt{202}. The overall trajectory highlights the strong continuity of the \textsc{EVE} backbone across training.

\subsubsection{Stage-1 search}

Table~\ref{tab:stage1_search_summary} reports the observed stage-1 search candidates. This stage explores a compact region around $\lambda_{\text{band,high}} \in \{2.00, 2.05, 2.10\}$.

\begin{table}[htbp]
\centering
\caption{Stage-1 EVE search summary on the new run.}
\label{tab:stage1_search_summary}
\begin{tabular}{lccccc}
\hline
Candidate & $\lambda_{\text{band,high}}$ & CE & Local & Frac.\ High & $\mu^2$ \\
\hline
Stage1 01 & 2.00 & 5.4111 & 0.1318 & 0.0856 & 0.0497 \\
Stage1 02 & 2.05 & 5.4094 & 0.1264 & 0.0821 & 0.0484 \\
Stage1 03 & 2.10 & 5.4381 & 0.1267 & 0.0855 & 0.0498 \\
\hline
\end{tabular}
\end{table}

At this short-budget stage, $\lambda_{\text{band,high}}=2.05$ gives the lowest CE. The setting $2.00$ remains very close. The useful search region is therefore compact and well centered around the low-$2.x$ range, which is a favorable outcome for controlled tuning.

\subsubsection{Extended uncertainty metrics}

Table~\ref{tab:uncertainty_summary_eve_det} reports the extended uncertainty and diversity summary. The predictive metrics are reported for both model families. The structural quantities follow the evaluation conventions defined in Section~\ref{sec:method}.

\begin{table}[htbp]
\centering
\caption{Extended uncertainty and diversity summary on the new run.}
\label{tab:uncertainty_summary_eve_det}
\begin{tabular}{lcccccccc}
\hline
Model & NLL & ECE & MI & Epi. & Flip & CVaR-NLL & Frac.\ High & $\mu^2$ \\
\hline
EVE & 5.6404 & 0.0326 & 0.1777 & 0.0325 & 0.3197 & 12.5035 & 0.0826 & 0.0487 \\
DET & 5.6698 & 0.0792 & 0.0000 & 0.0000 & 0.0000 & 13.0499 & 0.0000 & 0.7370 \\
\hline
\end{tabular}
\end{table}

Relative to \textsc{DET}, \textsc{EVE} improves NLL by $0.0294$, reduces ECE by $0.0466$, and improves CVaR-NLL by $0.5464$. The variational backbone also provides non-zero mutual information, non-zero epistemic spread, and a non-zero top-1 flip rate under Monte Carlo sampling. These results support the central claim that \textsc{EVE} is not only a stronger predictor, but also a richer carrier of usable uncertainty.

\subsubsection{Latent-state diagnostics}

\begin{figure}[htbp]
\centering
\includegraphics[width=0.9\linewidth]{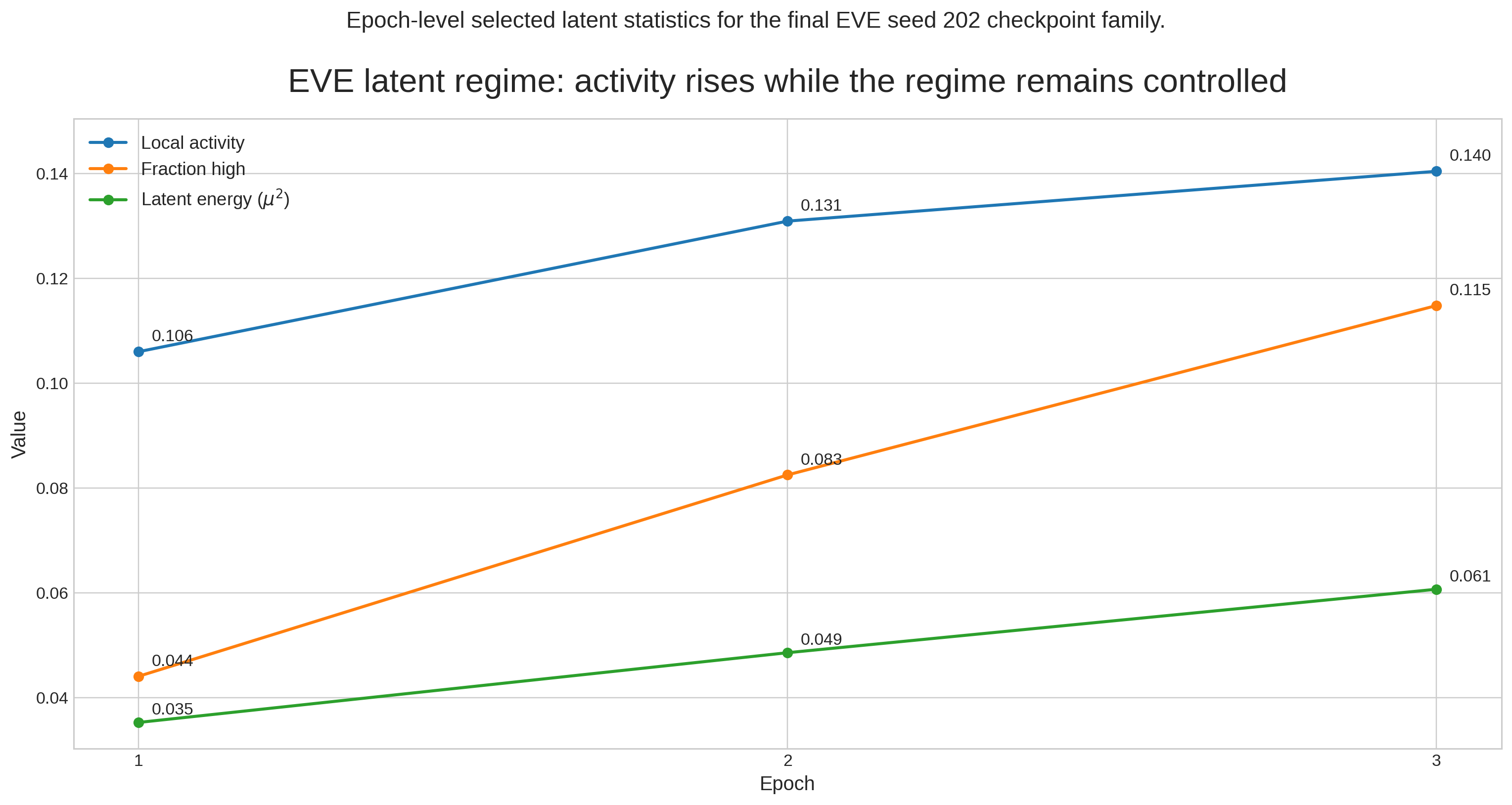}
\caption{
Epoch-level latent-state diagnostics for the retained \textsc{EVE} checkpoint family. The figure tracks local activity, the fraction of high-activity units, and latent energy across epochs.
}
\label{fig:eve_latent_regime}
\end{figure}

Figure~\ref{fig:eve_latent_regime} reports the epoch-level latent-state diagnostics for the retained \textsc{EVE} checkpoint family.

\FloatBarrier

The latent regime stays active, stable, and well controlled throughout training. In the final traces, local activity rises from about $0.10$ at epoch~1 to about $0.13$--$0.14$ by epoch~3. Over the same interval, the fraction of high-activity units moves from about $0.04$ to about $0.11$, and $\mu^2$ rises from about $0.035$ to about $0.06$. This is the expected profile of a live variational backbone with sustained internal organization.

\subsection{Full multi-action agentic evaluation}

The retained \textsc{EVE} model also supports a full multi-action agentic evaluation. This protocol uses the calibrated uncertainty controller with multiple support actions. The reported action rates follow the conventions defined in Section~\ref{sec:method}: coverage and abstention are terminal outcomes, while retrieve, deliberate, and resample are support-action activations and may overlap.

The controller uses the reported uncertainty thresholds
\[
uq_{\text{green}} = 0.6069,\qquad
uq_{\text{orange}} = 0.7500,\qquad
uq_{\text{red}} = 0.8919.
\]

\begin{table}[htbp]
\centering
\caption{Full agentic evaluation summary on the new run.}
\label{tab:agentic_full_eval_summary}
\begin{tabular}{lc}
\hline
Metric & Value \\
\hline
Selected seed & 101 \\
Number of examples & 200 \\
Accepted examples & 180 \\
Abstained examples & 20 \\
Coverage & 0.9000 \\
Accepted accuracy & 0.1778 \\
Overall accuracy & 0.1600 \\
Accepted CE & 4.6553 \\
Overall CE & 4.6553 \\
Mean cost & 2.7600 \\
Mean MC cost & 8.3800 \\
Mean steps & 1.7450 \\
Abstain rate & 0.1000 \\
Direct rate & 0.6350 \\
Retrieve rate & 0.3200 \\
Deliberate rate & 0.2650 \\
Resample rate & 0.2650 \\
Utility & 0.1048 \\
Avoided errors vs direct & 0.0994 \\
\hline
\end{tabular}
\end{table}

The controller answers $180$ of $200$ examples, for a coverage of $0.9000$. The accepted accuracy reaches $0.1778$, with overall accuracy $0.1600$. The mean cost is $2.7600$, the mean Monte Carlo cost is $8.3800$, and the reported utility is $0.1048$. Retrieval and deliberation are both actively used in this protocol, and the controller avoids a measurable fraction of direct-model errors. This gives a direct end-to-end demonstration that internal uncertainty signals can support useful downstream control.

\begin{figure}[htbp]
\centering
\includegraphics[width=0.9\linewidth]{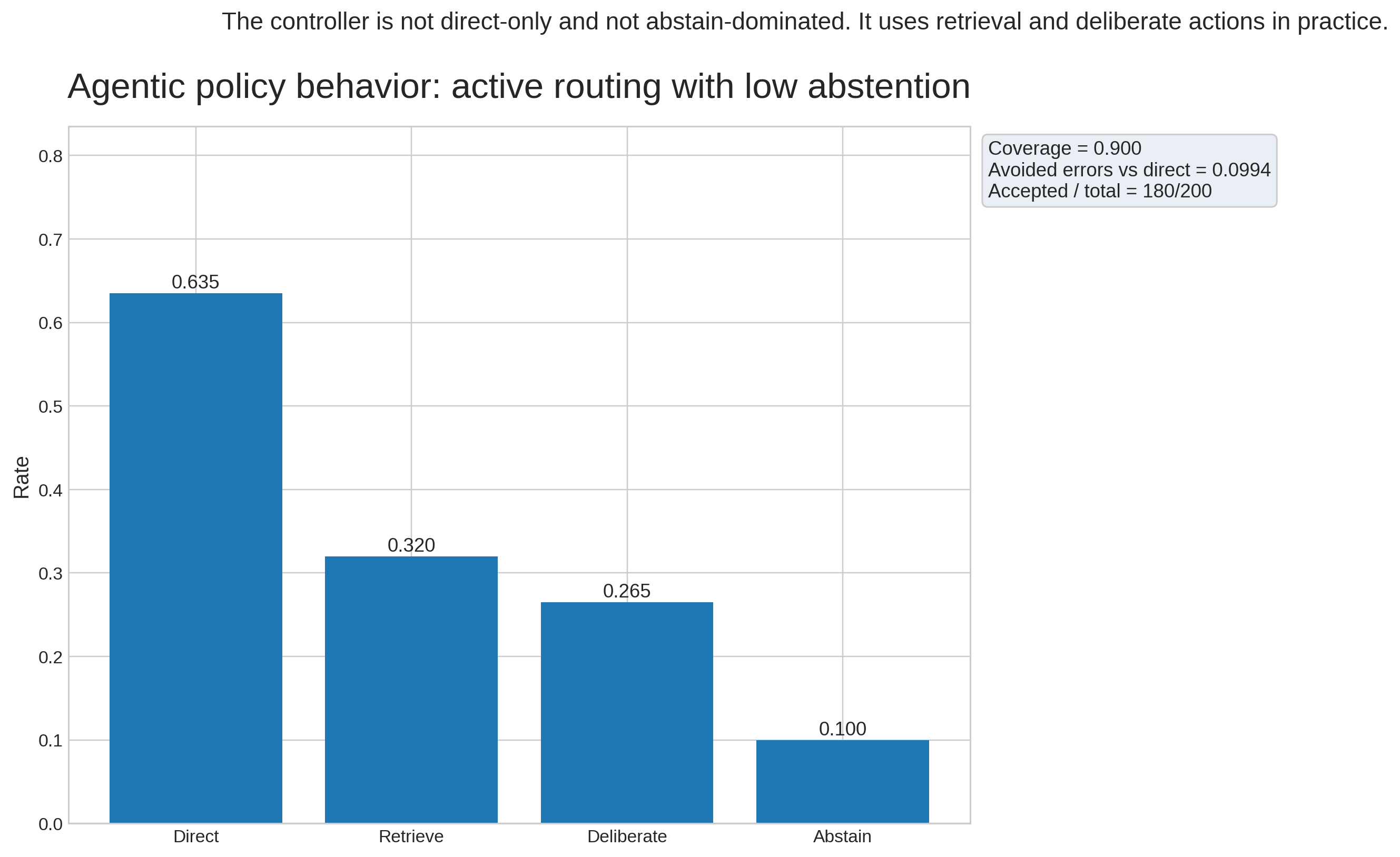}
\caption{
Observed behavior of the calibrated controller in the full multi-action evaluation. The summary box reports coverage, accepted/total examples, and avoided errors relative to direct prediction.
}
\label{fig:agentic_policy_behavior}
\end{figure}

\FloatBarrier

\subsubsection{Verification checks}

Table~\ref{tab:agentic_verification} reports the observed verification checks for the same run.

\begin{table}[htbp]
\centering
\caption{Agentic verification results on the new run.}
\label{tab:agentic_verification}
\begin{tabular}{lcc}
\hline
Criterion & Passed & Value \\
\hline
Coverage is sufficient & True & 0.9000 \\
Retrieve is used & True & 0.3200 \\
Deliberate is used & True & 0.2650 \\
Not abstain-dominated & True & 0.1000 \\
Utility is positive & True & 0.1048 \\
Avoids some direct errors & True & 0.0994 \\
\hline
\end{tabular}
\end{table}

All reported checks pass on this evaluation. The final run therefore supports a consistent conclusion across backbone quality, uncertainty quality, and downstream control utility. The variational backbone improves over the deterministic reference on the retained validation metrics, and the calibrated controller converts these internal uncertainty signals into a positive agentic outcome.

\subsection{Summary of observations}

Across the reported experiments, the retained \textsc{EVE} model improves on the matched \textsc{DET} reference on the main backbone metrics and exposes non-zero epistemic uncertainty signals. In the full multi-action setting, the calibrated controller remains active while preserving high coverage and positive utility. Together, these results provide the empirical basis for the discussion in the next section.

\section{Discussion}

The results establish a clear picture of the paper's central contribution.
They show that a variational language model can sustain a compact and effective form of internal agentic control.
In this setting, uncertainty becomes an operational signal that connects training-time regulation, checkpoint retention, and inference-time intervention.

A first result is that the variational backbone serves two valuable functions at once.
It remains strong on the base language-modeling task, and it exposes a richer uncertainty structure than the matched deterministic reference.
These two properties reinforce each other.
Uncertainty becomes most useful when it is carried by a backbone that also performs well as a predictor.
From this perspective, the results show that variational hidden computation can support both prediction and control within the same model.

A second result is the clear organization of regulation, retention, and routing into a single end-to-end pipeline.
The experiments give this structure a practical meaning.
The homeostatic regulator maintains the latent regime in a stable and useful range during training.
The retention rule preserves a checkpoint whose internal state remains well suited for later control.
The calibrated controller then acts on top of that retained state.
This sequence gives the full system strong internal coherence.
The controller operates on uncertainty that has already been stabilized, shaped, and preserved for decision-making.

The agentic evaluation sharpens this picture further.
In the full multi-action setting, the controller remains active while preserving broad coverage.
This shows that the richer uncertainty structure of the variational model can support a meaningful and flexible decision process under an explicit quality--cost trade-off.
The result is not only that uncertainty is measurable, but that it becomes operational in downstream control.

These results also make the role of the framework clearer within the broader landscape of agentic systems.
The paper establishes a minimal and measurable form of agentic control grounded in internal model evidence.
It connects naturally with the broader literature on retrieval agents, tool use, and environment-facing reasoning by contributing an earlier internal layer in the stack.
Here, the language model already supports regulation and action selection through its own internal evidence.
This makes the framework valuable as a building block for larger systems.
A model that sustains calibrated internal control can provide a strong substrate for later external agentic extensions.

The results point to several strong next steps.
One direction is scale.
It will be valuable to test the same methodology in larger and more varied language-modeling settings.
A second direction is policy design.
Richer routing policies could combine calibrated internal uncertainty with stronger retrieval or external support mechanisms.
A third direction is architectural breadth.
It will be useful to test whether the same separation between latent regulation, checkpoint retention, and calibrated routing remains effective beyond the present backbone.

Overall, the paper presents a coherent end-to-end picture in which internal stochastic evidence remains useful from training through inference.
That continuity gives the proposed notion of minimal agentic control its methodological strength.

\section{Conclusion}

This paper studied whether a variational language model can support a minimal and measurable form of agentic control grounded in its own internal evidence.
To address this question, we combined four elements:
local variational hidden computation, homeostatic latent regulation, structurally aware checkpoint retention, and a calibrated uncertainty-aware controller applied to the retained model.

The empirical results support a clear conclusion.
The variational backbone improves over a matched deterministic reference on the language-modeling task while also exposing a richer uncertainty profile.
On top of this backbone, the calibrated controller uses internal uncertainty to guide inference-time decisions in a full multi-action setting.
Across the reported results, the same internal evidence supports both model characterization and action selection.

The main conclusion is precise.
Internal uncertainty in a variational language model can function as a practical control interface.
It can regulate the latent regime during training, support the retention of a structurally meaningful model state, and guide calibrated intervention at inference time.
This yields a compact and concrete form of agentic control inside the model itself.

More broadly, the paper shows that agentic behavior can begin inside the model through the disciplined use of internal evidence.
From there, it can be extended through tools, environments, and external orchestration.
In that sense, this work provides a strong foundation for future systems that combine internal uncertainty and external action within a larger agentic architecture.

\bibliographystyle{plainnat}
\bibliography{references}

\end{document}